\relax
\documentclass{article}
\usepackage{ijcai17}
\usepackage{times}
\usepackage{helvet}
\usepackage{courier}
\usepackage{amsmath,amsfonts,amssymb,bm}
\usepackage{varwidth}
\usepackage{xcolor}
\usepackage{url}
\usepackage{booktabs} 
\usepackage{array}
\usepackage{multirow}
\usepackage{graphicx}
\usepackage{amsmath}
\usepackage{stmaryrd}
\usepackage[small]{caption}

\newcommand\citeA[1]{%
  \citeauthor{#1} [\citeyear{#1}]}

\begin{document}
%
\title{Predicting the Quality of Short Narratives from Social Media}
\author{Tong Wang$^{1,2}$, Ping Chen$^{1}$, and Boyang Li$^2$\\
    $^1$University of Massachusetts Boston\\
    $^2$Disney Research\\
	\{twang, ping.chen\}@cs.umb.edu, albert.li@disneyresearch.com\\
}
\maketitle
\begin{abstract}
An important and difficult challenge in building computational models for narratives is the automatic evaluation of narrative quality. Quality evaluation connects narrative understanding and generation as generation systems need to evaluate their own products. To circumvent difficulties in acquiring annotations, we employ upvotes in social media as an approximate measure for story quality. We collected 54,484 answers from a crowd-powered question-and-answer website, Quora, and then used active learning to build a classifier that labeled 28,320 answers as stories. To predict the number of upvotes without the use of social network features, we create neural networks that model textual regions and the interdependence among regions, which serve as strong benchmarks for future research. To our best knowledge, this is the first large-scale study for automatic evaluation of narrative quality. 
\end{abstract}

\section{Introduction}
The quest for an Artificial Intelligence (AI) that can process narrative information dates back from early years of AI research. 
Recently we have seen renewed interest in narrative generation \cite{Koncel2016,Li2013} and narrative understanding \cite{Bamman2013,Ouyang2015,Pichotta2016,Ferraro2016,Srivastava2016}. An important and difficult problem that is shared by both lines of research is the evaluation of narrative quality. Naturally, being able to evaluate the quality of a narrative requires a sufficient level of narrative understanding. In addition, the ability to evaluate its own product is vital for a narrative generation system. In this paper, we present the first large-scale attempt to tackle this challenging problem. 

The first challenge is the lack of human annotation. Aesthetic evaluation of a complex artifact like a story is inherently subjective, so obtaining a reliable majority opinion may require a large number of human judges and may be expensive. To circumvent this issue, we turn to narrative content from social media and use the number of upvotes they receive as a proxy measure for quality. In doing so, we aim to predict the reaction of a large population to a story rather than to simulate the professional judgment of  literary critics. 

Specifically, we utilized a popular question-and-answer website, Quora, where users can ask questions, write answers, and cast upvotes or downvotes on the answers.  
Answers to questions like ``what does it feel like to be poor'' are likely stories. A question on Quora is usually labeled with a few topics by the crowd.
We manually identified 21 Quora topics that contain questions whose answers are likely stories. We created a classifier to separate story-like answers from non-stories using active learning, yielding 28,320 story texts under 742 questions. 

\begin{table}[t]
  \centering
  \caption{A story with complex interaction among events.}
 {\begin{tabular}{c|l}
    \toprule     	
     1 & Alice wired 20 million into Bob's offshore account. \\
	 2	& One night, Bob murdered Alice's husband. \\
	 3	& A weekly later, Alice murdered Bob. \\
    \bottomrule
  \end{tabular}}
  \label{murder-story}
\end{table}

The second challenge is to understand complex semantics of narratives. Quora stories tend to be short and use simple language, which simplifies but does not completely eliminate the problem. A common computational model for a story is a sequence of events that interact and collectively reveal higher-level semantic constructs like character intentions \cite{Ferraro2016,IPOCL10}. An example story in Table \ref{murder-story} illustrates the interaction among events. As presented, the story is about a double cross, but its significance can be completely changed by removing either event 1 or event 3. Removing event 1 leaves a revenge story; removing event 3 turns it into a mercenary story. Based on the intuition that (1) a story contains multiple textual chunks and (2) high-level semantic meaning is expressed by the interplay of textual chunks, we propose neural networks that learn representation for textual regions and weigh the regions according to their interdependence. 

%


In order to capture the effects of content, we refrain from using social network features in the prediction. 
Studies on social networks suggest that the initial reaction to content (e.g., the time it takes to get the first 100 retweets) or network structures are strong predictors for popularity in social media~\cite{Tatar2014,Leskovec2014}. 
To an extent, these features are surrogates for story understanding. 
For instance, the first batch of human readers may work as filters who recognize high-quality content. A large number of followers may be due to the author's ability to consistently produce good content.
Hence, these features can be considered as a form of human computation. 
Our goal in this paper is not to use human surrogates and to build purely computational models for story understanding. In doing so, we hope to improve content predictors as much as possible and generalize to evaluating stories outside social media. 



The contributions of this paper are two-fold. First, we introduce the use of upvotes on social media as a proxy measure for narrative quality and create a dataset for narrative quality evaluation. Second, we establish strong benchmarks for the quality evaluation task using neural models based on regional embeddings and their interdependence. Our best model achieves a 18.10\% reduction in mean square error relative to a strong random forest baseline. 

\section{Related Work}

To our knowledge, this paper represents the first \emph{large-scale} study dedicated to automatic quality evaluation of textual stories. Previous studies concern only a small number of stories or the products of one story generation systems. 
\citeauthor{McIntyre2009} \shortcite{McIntyre2009} studied the interestingness of 40 Aesop's fables. \citeauthor{Ganguly2014} \shortcite{Ganguly2014} argued that highlight by Amazon Kindle users is associated with aesthetics and predicted if a text segment from 50 English fictions is highlighted. 
\citeA{Milli2016} studied fanfictions and built predictors for readers' responses to characters in them. 
The story writing companion SayAnything \cite{SayAnything:Journal} selects coherent responses to user-generated content. Aesthetic measures, such as suspense \cite{Cheong:Suspenser,ONeill2014}, have also been built from symbolic representations of stories. 

More broadly, 
a number of works addressed general writing quality, including readability \cite{Pitler2008,Feng2010} and coherence \cite{Barzilay2008}. Though readability and coherence are important for story quality, story quality is a broader notion that are affected by other factors such as story structure \cite{Ouyang2015}. Kao and Jurafsky \shortcite{Kao2012} predicted if a poem is written by a professional poet or an amateur poet using sound device, word frequency, affect and imagery. 

A popular application of text quality evaluation is automatic grading of student essays \cite{Taghipour2016,Alikaniotis2016,Cummins2016}. However, those work usually focus on essays that aim to argue for one or two predefined viewpoints as part of a English language test, rather than creative writing or storytelling. Due to the standard nature of the language tests, there are usually a large number of student responses to a small number of questions. For example, the Automated Student Assessment Prize dataset contains 21,450 essays for 8 questions. Stories are usually much more diverse. The Quora dataset we collected contains 28,320 texts in answer to 742 questions.  

In social media research, many works addressed prediction of content popularity as well as the possibility of sharing cascades (i.e., content ``going viral''). Due to space constraints, we refer readers to the survey by \citeA{Tatar2014}.  \citeauthor{Leskovec2014} \shortcite{Leskovec2014} found initial popularlity to be highly predictive of cascades and content features to be weak predictors.
%
%
This paper differs from the social media perspective as we do not use features like initial popularity, which can only be observed once the content appears on social media. We also avoid features derived from social network structures or time series. Instead, the focus of this paper is to improve pure content predictors as much as possible. 


\section{Data Collection}
In this section, we describe our effort to collect the story dataset and describe its characteristics. 

Quora is a social question-and-answer website, where users can ask questions, write answers, and cast upvotes or downvotes on the answers. Some questions, such as ``what does it feel like to be poor'', ask directly for personal anecdotes and stories. 
Each question is labeled with one or more Quora-curated topics by the crowd. For example, the two questions above are both labeled with the topic \texttt{Survey Question}. Quora uses upvotes and downvotes as indicators of quality in recommending answers to users; highly rated answers are displayed first under a question. 

\begin{table}[t]
  \centering
  \caption{Statistics of the five Quora topics with the most answers collected.}
 {\begin{tabular}{l|c|c|c}
    \toprule 
    	Topic & \begin{tabular}[x]{@{}c@{}}Questions\\Collected\end{tabular} & \begin{tabular}[x]{@{}c@{}}Answers\\Collected\end{tabular}  & Stories\\
    	\midrule 
     Survey Question 		&73  &8,066 &4,370 \\
	 Experiences in Life	&68  &7,836 &4,321 \\
	 Life and Living 	    &80  &7,162 &3,895 \\
     Short Stories		    &75  &6,894 &3,853 \\
	 Life Lessons		    &68  &5,178 &2,837\\
    \bottomrule
  \end{tabular}}
  \label{quora-stats}
\end{table}

To bootstrap the data collection, we manually identified 21 Quora topics that likely contain questions with story answers. We collected all available questions shown on the topic page and answers under those questions, resulting in 54,484 answer texts. The number of upvotes and views were recorded. Images, if any, were discarded. Although Quora allows users to downvote an answer, the number is not displayed on the webpage and cannot be recorded. If an answer received more than 1,000 upvotes, Quora rounds the number to the hundreds and displays it like ``2.2k''. This introduced slight imprecision in the collected data. Table \ref{quora-stats} shows the five topics with the most collected stories.

\subsection{Separating Stories from Non-Stories}

Not all collected texts are stories. Common non-story texts include personal opinions on places and cultures, general observations, and personal advice. We employed an active learning approach to build a classifier that separates stories from non-stories. 
We randomly selected 1,000 texts that have a minimum of 50 words to be annotated on Amazon Mechanical Turk (AMT). In agreement with narratological theories \cite{Prince1987}, the AMT workers were instructed that a story is a sequence of events that are causally and temporally related. Having a specific time, location and conflict makes the story more prototypical, although each element is not strictly necessary. 
Three AMT workers read one text and classified it as either a story or a non-story. The three workers reached unanimous agreement on 68.8\% of the cases. The majority vote is used when an unanimous agreement is not available. In total, 637 out of the 1,000 texts are labeled as stories and 363 are labeled as non-stories.

For automatic labeling, we used logistic classification with distributed bag-of-words (DBoW) features. Formally, given a vocabulary $\mathcal{V}$, each word $w \in \mathcal{V}$ is mapped to an $D$-dimensional embedding $\pmb v_w$. Let $y_i$ denote the label for a text $x_i$, and $\pmb \theta$ denote model parameters. The probability that $x_i$ is a story is computed as
\begin{equation*}
P(y_i=1|x_i) = \frac{1} {1+ \exp \pmb \theta^{\top} \pmb u} \, , \,\, \pmb u = \sum_{w \in x_i} \pmb v_w 
\end{equation*}
%
The model was trained to maximize the probability of correct labels in the training set. 
We obtained 300-dimensional word embeddings using the \texttt{word2vec} algorithm \cite{Mikolov2013} from 4 gigabytes of textual data, which contain fictions from Project Gutenberg and plot summaries of movies and books from Wikipedia. 
The logistic classifier achieved 0.875 for precision,  0.937 for recall, and 0.905 for F1. 

We used an active learning approach to further improve accuracy on difficult texts. We built a committee of 11 classifiers using bagging (achieving 0.907 for F1), and selected 400 additional texts that the committee had the least agreement on. These 400 texts were sent to AMT for a second round of labeling. These 400 texts appeared to be difficult for human annotators as well, as the proportion of unanimous agreement among three AMT workers fell to 51.3\%. When we trained the same logistic classification model using 1,000 texts from the first batch and 200 texts from the second batch, the prediction performance on the remaining 200 difficult texts dropped to 0.572 for F1.

 \begin{table}[t]
  \centering
  \caption{Performance of story classification on all 1,400 annotated texts.}
 {\begin{tabular}{l|c|c|c}
    \toprule 
    	Classifier & Precision & Recall & F1 \\
    	\midrule 
    LR + BoW 				& 0.790 & 0.916 & 0.847 \\
    LR + TF-IDF 				& 0.593 & 1.0 & 0.744 \\
	LR + DBoW	 & 0.775 & 0.869 & 0.815\\
	RF + BoW  	& 0.773 & 0.946 & \textbf{0.850} \\
   RF + DBoW  	& 0.804 &  0.881 & 0.840 \\
   SVM + BoW & 0.832 &  0.818 & 0.827 \\
   SVM + DBoW & 0.799 &  0.868 & 0.832 \\
    \bottomrule
  \end{tabular}}
  \label{tab:story-classification}
\end{table} 

With all 1,400 annotated texts, we experimented with a number of classification techniques, including logistic regression (LR), random forest with 500 trees (RF), and linear support vector machine (SVM). The features we use include bag-of-Words (BoW) frequency, term-frequency inverse-document-frequency (TF-IDF) features, and DBoW features. The BoW frequency features are calculated as a $|\mathcal{V}|$-dimensional vector $\pmb b$, whose $j^\text{th}$ component $\pmb b_j$ is the equal to the count of the word $w_j$ in the text. The TF-IDF features replace the word counts with the TF-IDF of the word $w_j$. 
Table \ref{tab:story-classification} summarizes the performance of different models as evaluated with 10-fold validation.
A random forest with BoW features achieves the best F1 of 0.850. 

We used the best RF classifier to classify all 54,484 answers, yielding 29,846 stories and 24,638 non-stories. Since a text with too few words are probably not interesting, we removed stories with less than 50 words. Furthermore, to increase the reliability of upvote data, we removed stories with less than 50 views. This procedure obtained 28,320 stories for upvote prediction.

\begin{table}[t]
  \centering
  \caption{Negative log-likelihood (NLL) of data for distributions fitted with MLE and the number of parameters fitted.}
 {\begin{tabular}{l|c|c}
    \toprule 
    	Distribution & NLL & \# Parameters \\
    	\midrule     
    Pareto 				& \textbf{133514} & 1 \\
    Log-normal  	& 144765 & 2 \\
    Exponential				& 187594 &  1 \\
	Half-normal	 & 221202 & 1\\ 
	Pareto + Exponential  	& \textbf{132848} & 2 \\	  
    Gaussian & 238675 &  2\\
    5 Gaussians & 139462 & 10 \\
    \bottomrule
  \end{tabular}}
  \label{tab:distribution-fit}
\end{table} 

\begin{figure}[t]
  \caption{The Pareto distribution fitted using MLE on log-log scale. The dashed straight line shows the fitted Pareto distribution and the green dots represent observed data.}
  \centering
    \includegraphics[width=0.45\textwidth]{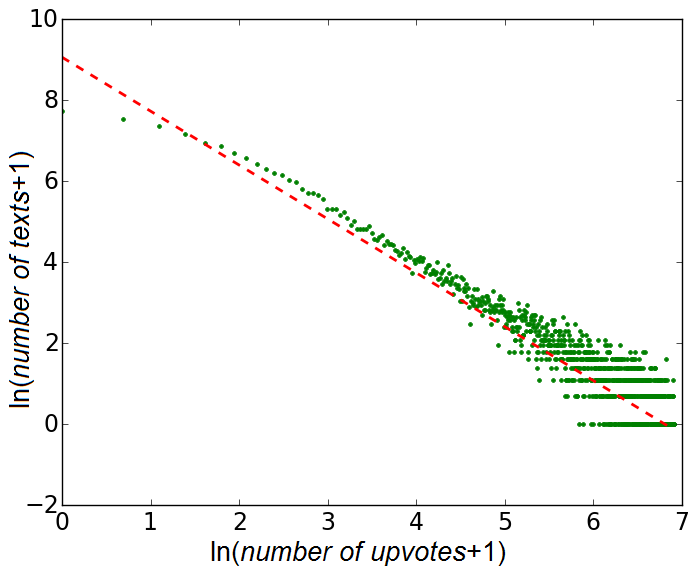}
  \label{fig:loglog}
\end{figure}

\subsection{Statistical and Linguistic Properties}

As we aim to predict the number of upvotes, we set out to characterize its probabilistic distribution, which is highly skewed with a heavy tail, with a mean of 453.8 and a median of 11. The standard deviation is 2215.7. The 10\% and 90\% quantiles are 1 and 379 respectively. We fitted several distributions using the maximum likelihood estimator (MLE). Table~\ref{tab:distribution-fit} summarizes the goodness of fit for the following distributions: Pareto, exponential, lognormal, half-normal, the mixture of Pareto and exponential, and mixtures of Gaussian distributions. For fitting the Pareto distribution with the support $[x_m, +\infty]$, we set $x_m$ to 1. For both Pareto and log-normal, we add 1 to all upvote counts before MLE. 
The mixture of Pareto and exponential provides the best fit and the best Bayesian information criterion. Pareto is a close second and makes use of only one parameter. Log-normal also provides a decent fit. Figure \ref{fig:loglog} shows the fitted Pareto distribution on log-log chart. 

Compared to complex literary stories, most Quora stories are short and informal. The average length of stories is 369.3 words with a standard deviation of 288.9. The median length is 288. The 10\% and 90\% quantiles are respectively 123 and 705. The average Glesch-Kincard Grade Level, computed with Flesh tool\footnote{\url{http://flesh.sourceforge.net/}}, is 8.32, putting the stories around secondary school reading level. 

\section{Predicting Story Quality}
\begin{figure*}[t]
  \caption{Block diagrams for three neural network models: (a) the regional story reader, (b) the sequential reader, and (c) the holistic reader. Due to space concerns, the dotted boxes (``Regional'') in (b) and (c) are equivalent to the regional components in (a), which include the convolution layers, the pooling layers and one fully connected layer. Repeated blocks are shown with multipliers like ``$2\times$'' and ``$3\times$'' on dashed boxes.}
  \vspace{0.05in}
  \centering
    \includegraphics[width=\textwidth]{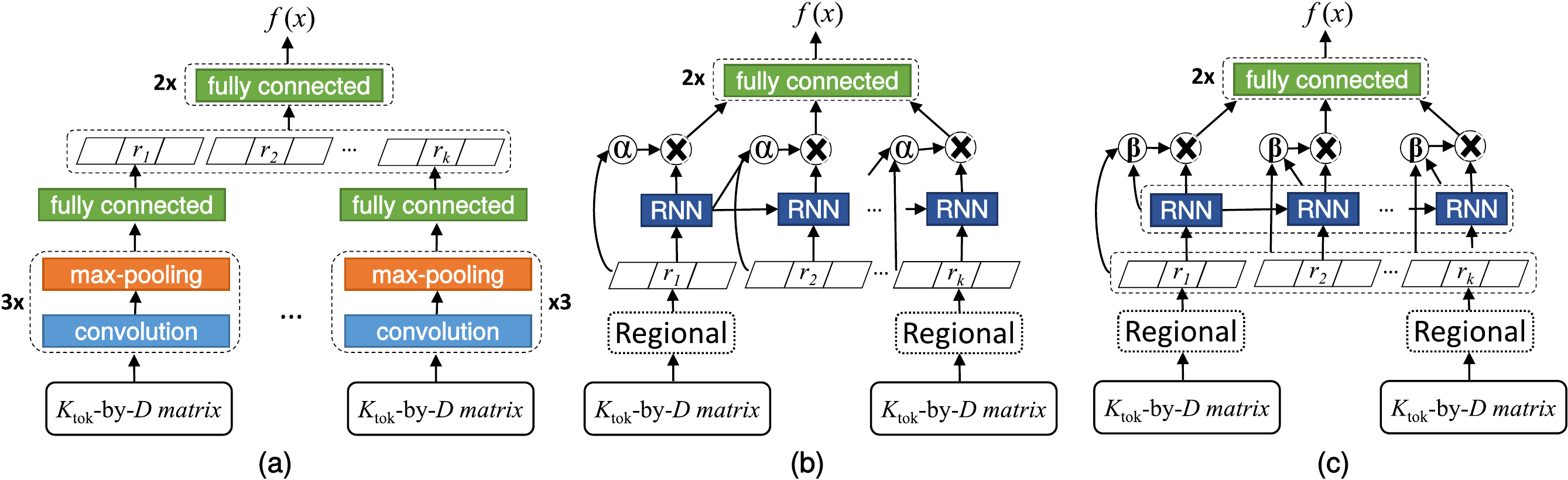}
  \label{fig:networks}
\end{figure*}

In this section, we describe convolutional neural networks (CNN) for predicting the upvotes received by texts, starting with a simple regional model, followed by two variants that further capture interdependence among regions. Figure \ref{fig:networks} illustrates the three network architectures. 

\subsection{A Regional Story Reader}
In a regional CNN model, we divide the entire text into $k_{reg}$ regions, each with $k_{tok}$ words. Every word is associated with a precomputed embedding in a $D$-dimensional space. Thus, we can represent a region by arranging the word embeddings into a $k_{tok}$-by-$D$ matrix. We apply three convolution layers with the linear rectifier (ReLU) activation function, each followed by a max-pooling layer. 
The convolutional filters are shared across all regions. After that, we connect the output to one fully connected layer to reduce its dimensionality to $D^r$, and we denote its output as $\pmb r_t$ for the $t^{\text{th}}$ region. The document representation $\pmb d^r$ is the concatenation of all regional embeddings.
\begin{equation}
\pmb d^r = \begin{pmatrix} \pmb r_1^\top & \pmb r_2^\top & \cdots & \pmb r_{k_{reg}}^\top \end{pmatrix}^\top
\end{equation}
The network further contains two fully connected layers with ReLU activation on top of $\pmb d^r$. Dropout layers with 50\% dropout rate are inserted after the CNN and between any two fully connected layers. 

\subsection{A Sequential Reader}
The sequential story reader extends the regional reader by introducing recurrent and gating units that represent the interdependence between different regions. A recurrent neural network (RNN) connects regions through the story. Let $\pmb h_t$ denote the hidden state of the RNN at time step $t$, $\pmb W^{h}$, $\pmb U^{h}$, $\pmb b^h$, $\pmb W^{\alpha}$, $\pmb U^{\alpha}$, and $\pmb b^\alpha$ denote matrices and vectors that contain trainable parameters. We have the following recurrence equation:
\begin{equation}
\label{eq:rnn}
\pmb h_t = \text{tanh}(\pmb W^{h} \pmb r_t + \pmb U^{h} \pmb h_{t-1} + \pmb b^h)
\end{equation}
We introduce a gate vector $\pmb \alpha_t$, which attenuates the regional embedding depending on previous regions. $\pmb \alpha_t$ is of the same dimension as $\pmb r_t$. It is computed using the previous hidden state $\pmb h_{t-1}$ and the current regional embedding $\pmb r_t$.
\begin{equation}
\label{eq:gate}
\pmb \alpha_t = \sigma(\pmb W^{\alpha} \pmb r_t + \pmb U^{\alpha} \pmb h_{t-1} + \pmb b^\alpha)
\end{equation} 
The final regional representation $\pmb s_t$ is computed as component-wise multiplication (denoted by $\otimes$).
\begin{equation}
\pmb s_t = \pmb \alpha_t \otimes \pmb h_t
\end{equation}
The document embedding $\pmb d^s$ is the concatenation of $\pmb s_t$.
\begin{equation}
\pmb d^s = \begin{pmatrix} \pmb s_1^\top & \pmb s_2^\top & \cdots & \pmb s_{k_{reg}}^\top \end{pmatrix}^\top
\end{equation}
In the above, . tanh$(\cdot)$ is the component-wise hyperbolic tangent function and  $\sigma(\cdot)$ is the component-wise sigmoid function. Similar to the regional reader, we connect $\pmb d^s$ to two fully connected layers with the ReLU activation function. The same dropout layers are used as in the regional reader. 

Note the gate in Eq. \ref{eq:gate} is the same with the output gate in long short-term memory (LSTM) \cite{Hochreiter1997}. The above singly gated RNN can be seen as a compromise between RNN and LSTM. We may replace this RNN with a full-fledged LSTM recurrence between regions.

\subsection{A Holistic Reader}
The holistic story reader is similar to the sequential reader except it allows all other regions to influence the gates on the current region. The intuition behind this architecture is that the meaning of individual events and regions emerge from entire story. 

The RNN network described by Eq. \ref{eq:rnn} remains unchanged. The computation of the gate vector, which is now denoted by $\pmb \beta_t$, is changed to the following, with network parameters $\pmb w^{\beta}_t$, $\pmb u^{\beta}_t$ and $\pmb b^\beta$. 
\begin{equation}
\label{eq:holistic-gate}
\pmb \beta_t = \sigma( \begin{pmatrix} \pmb r_1 \\ \pmb r_2 \\ \cdots \\ \pmb r_{k_{reg}} \end{pmatrix}^\top \pmb w^{\beta}_t + \begin{pmatrix} \pmb h_1 \\ \pmb h_2 \\ \cdots \\ \pmb h_{k_{reg}} \end{pmatrix}^\top  \pmb u^{\beta}_t + \pmb b^\beta)
\end{equation}
The computation of regional and document ($\pmb d^h$) representations remains the same.
\begin{equation}
\pmb s_t = \pmb \beta_t \otimes \pmb h_t
\end{equation}
\begin{equation}
\pmb d^h = \begin{pmatrix} \pmb s_1^\top & \pmb s_2^\top & \cdots & \pmb s_{k_{reg}}^\top \end{pmatrix}^\top
\end{equation}
Two fully connected layers on top of $\pmb d^h$ as well as the dropout layers remain unchanged from the Sequential Reader. We can also replace the RNN with LSTM, whose outputs go through the holistic gating mechanism described in Eq. \ref{eq:holistic-gate}.

\subsection{Other Features}
Besides the word embeddings, we employ 26 linguistic features that are correlated with the aesthetic quality of text. 
The ability to portray a clear mental image is likely an indication of good stories. In poetry evaluation \cite{Kao2012} and fairy tale generation \cite{McIntyre2009}, the number of words that refer to concrete objects was used as a proxy measure for imagery. In this paper, we employ the expert-designed word categories from Harvard General Inquirer (HGI) \cite{Stone1966}, and count how many words in each answer are classified under the Object category. 
In addition to imagery, we also selected 20 other HGI categories, such as positive and negative words, as additional features. 
Moreover, we extract the following features from the answer: the number of times the answer was viewed, the logarithm of the views, the number of images, words, and grammatical mistakes. These features are used in all models we created. In the neural network models, they are fed into the earliest fully connected layer. 

\subsection{Loss Function}
As shown earlier, the distribution of upvotes can be described by heavy-tailed distributions like the Pareto and the log-normal distribution. For simplicity, we optimize for log-likelihood under the log-normal distribution. Let $x_i$ denote the features representing the $i^\text{th}$ text and $y_i$ its number of upvotes. The neural network function is $f(x_i; \pmb \theta)$ where $\pmb \theta$ are the parameters. 
Under the log-normal distribution, the probability density function parameterized by the mean $\mu$ and standard deviation $\sigma$ is 
\begin{equation}
P(\pmb y|\mu, \sigma) = \prod_i \frac{1}{\sigma \sqrt{2\pi}}  \exp\frac{-(\ln y_i - \mu)^2}{2 \sigma^2}
\end{equation}
Setting $\hat \mu_i = f(x_i; \pmb \theta)$ and holding $\sigma$ constant, we can easily show that minimizing the following square loss function is equivalent to maximizing the log-likelihood of observed data.
\begin{equation}
\mathcal{L}(\pmb y, \pmb x, \pmb \theta) = \sum_i (\ln y_i - \hat \mu_i)^2
\end{equation}
We use this loss function for all models in the experiments. To avoid the undefined $\log 0$, we add 1 to all upvote counts. 

\subsection{Dividing Texts into Regions}
The three reader models divide the answer text into $k_{reg}$ regions. As the question may provide additional information, we put the question text as the first region with zero padding, yielding $k_{reg} + 1$ regions in total. The convolutional components share the same weights across all answer regions, but the question region employs its own set of weights. This weight-sharing configuration is motivated by the linguistic and semantic differences in question and answer texts. 

The region size is fixed to $k_{tok}$. Thus, the reader networks are built on exactly $k_{reg} \times k_{tok}$ words, but the answer text may contain more or less words. To create a relatively faithful representation of the text, we evenly divide the answer text into $k_{reg}$ partitions and take the first $k_{tok}$ words from the middle of every partition. If a partition is shorter than $k_{tok}$ words, zero paddings are used. 




\section{Experiments}
This section describes our experiment to test the three models' abilities to predict the number of upvotes received stories. 
All of the convolutional networks use the same setup. The first layer contains 32 3-by-5 filters with a horizontal stride of 3, followed by 32 2-by-3 filters with a horizontal stride of 2, and 16 1-by-3 filters with strides of 1. Each filter layer is followed by a ReLU activation function and max-pooling layers with 2-by-2 kernels. The two fully connected layers applied to the document embedding contain 128 units each. As the average length of stories is 369.3, we set the region size $k_{tok}$ to 36 and the number of regions $k_{reg}$ to 10. The question corresponding to the answer is put into the first region, yielding 11 regions in total. Zero padding is used for stories that do not fill an entire region. Thus, the embedding of an empty region is an all-zero vector. The dimension of regional embeddings $\pmb r_t$ is set to 10. Thus, $\pmb \alpha_i$ and $\pmb \beta_i$ are of the same dimension, and the document embedding has 110 dimensions. As noted earlier, the sequential reader and the holistic reader can utilize either RNN or LSTM recurrence between the regions, so we use both variants for each reader.
We use the 300 dimensional pre-trained word embeddings from 4GB of story text using the \texttt{word2vec} technique, as we did previously. Stop words are set to a single randomly initialized vector. The word embeddings are fixed during training.

We partition the 28,320 story texts into a training set of 21,230 (75\%) stories, a validation set of 2,832 (10\%) stories and a test set of 4,258 (15\%) stories.  We tune hyperparameters and find the best model in 100 epochs using the validation set. We report the mean square error (MSE) on the test set.


We create the following baselines for comparison.
\noindent \textbf{Random forest.} The random forest (RF) outperformed all other classifiers in the story classification task. In the upvote prediction task, we create a random forest with 1,000 trees. The features used by the RF regressor include BoW word counts and all features available to the neural networks except the word embeddings. This is because we achieved better performance using pure word counts with random forest than using word embeddings in the story classification task. 

\noindent \textbf{Lasso and SVM.} In addition, we also use a lasso regression and a support vector machine with a radial basis function (SVM-RBF) kernel as two traditional baselines. The regularization constant for the lasso regression is set to 0.01. 

\noindent \textbf{Recurrent neural networks.} We create RNN and LSTM baselines (RNN-A and LSTM-A) as they are popular choices for textual data. Average-pooling over all hidden states is used to create a document embedding $\frac{1}{m}\sum_{t=1}^m \pmb h_t$, as we found this to outperform several other pooling methods. Similar to the regional model, the maximum number of words, $m$, is set to 360. Zero paddings are used for shorter text. Two fully connected layers of 128 units are applied subsequently. The dimension of hidden states is set to 100; the magnitude of gradients is clipped at 1. 

\subsection{Results and Discussion}
Table \ref{tab:upvote-prediction} shows the performance of all models as measured by mean square error. The random forest is still the strongest traditional technique, beating Lasso and SVM by large margins. But it is inferior to the neural techniques for this difficult task. The weakest neural baseline, RNN, outperforms the random forest by relative 9.76\%. 

The three reader models we introduced are shown to be the best models. In particular, the holistic reader paired with LSTM recurrence between regions achieves the best performance, an MSE of 0.4438. This represents 18.10\% relative improvement over the RF and 3.96\% relative improvement over the regional reader, the best model without explicit recurrence between regions. Overall, the results suggest the benefits of modeling the interdependence between text regions for story understanding. 

In line with many findings in the literature, we find that LSTM generally performs better than RNN. 
Interestingly, the sequential reader with a singly gated RNN outperforms the same model with LSTM. Our interpretation is that, in the sequential reader setting, using only the output gate is more effective than using all gates of the LSTM.

An ablation study, shown in the second half of Table \ref{tab:upvote-prediction}, suggests the question plays a large role in the prediction. When we use only the 26 other features without the question or answer texts, the holistic reader is reduced to two hidden layers of 128 units (denoted by Holistic+26), which performs worse than the random forest with the 26 features and the answer text (RF+26+A). After the answer text is added to the holistic reader (Holistic+26+A), it performs on par with the RF baseline that uses all features and both question and answer. Adding the question text (Holistic+LSTM) leads to a further 18.10\% error reduction. We offer the hypothesis that it is easier to learn from the question than the answer due to the large variations in writing styles and skills exhibited by different Quora authors. In comparison, questions are usually written in simple language with few artistic expressions like metaphors or sarcasm. 


 \begin{table}[t]
  \centering
  \caption{Performance of upvote prediction, measured in mean square errors (MSE) and relative improvements over RF baselines.}
 {\begin{tabular}{l|l|l}
    \toprule 
    	Model & MSE & \% Improvement \\
    	\midrule 
    RF 				& 0.5419   & 0.00\% (baseline) \\
    Lasso				& 0.5953 &  -9.85\%\\
    SVM-RBF				& 1.2811 & -136.4\%\\
    RNN-A 				& 0.4895 & 9.67\%  \\
	 LSTM-A	         & 0.4776 & 11.87\% \\
	Regional  	   & 0.4621 & 14.73\%  \\
   Sequential+RNN  	& 0.4442 &  18.03\%  \\
   Sequential+LSTM & 0.4506 &  16.85\%\\
   Holistic+RNN & 0.4472  &  17.48\% \\
   Holistic+LSTM & \textbf{0.4438} & 18.10\% \\
   \midrule
   RF+26+A & 0.5563 & 0.00\% (baseline) \\
   Holistic+26 & 0.5758 & -3.51\%\\
   Holistic+26+A & 0.5420 & 2.57\% \\
    \bottomrule
  \end{tabular}}
  \label{tab:upvote-prediction}
\end{table}

\section{Conclusions and Limitation}
Quality evaluation of narratives provides a bridge between research on story generation and story understanding. In this paper, we built a dataset for the quality evaluation task by extracting stories and the upvotes they received from a social media website, Quora. We proposed several neural networks that model the textual chunks in a story and their interrelations, which achieves  18.10\% relative improvement over a random forest baseline and 3.96\% relative improvement over the best neural method without regional interdependence.  

Depsite some initial positive results on a difficult task, we note that upvotes in social media may be affected by factors other than quality, such as Quora's ranking function. Further research is needed to identify and correct such bias caused by social media. Narratives are complex and nuanced artifacts; an intermediate representation such as event constituents \cite{Luan2016} may further facilitate story understanding and quality evaluation. Although purely automatic evaluation of story generation systems remains an open problem, we believe the proposed networks will serve as strong baselines for future research. 

\section*{Acknowledgements}
Tong Wang was partially supported by the Oracle Doctoral Research Fellowship at the University of Massachusetts Boston.


\bibliographystyle{named}
\bibliography{references}
\end{document}